\documentclass[letterpaper,10 pt,conference]{IEEEtran}
\IEEEoverridecommandlockouts                             

\usepackage{float}
\usepackage{amsmath}
\usepackage{amssymb}
\usepackage{graphicx}
\usepackage{graphics}
\usepackage{blkarray}
\usepackage{booktabs}
\usepackage[pdftex,dvipsnames]{xcolor}  %
\usepackage{multirow}
\usepackage{comment}
\usepackage{epstopdf}
\usepackage{tikz}
\usetikzlibrary{arrows,matrix,positioning}
\usepackage[labelformat=simple]{subcaption}
\usepackage{soul}
\usepackage{MnSymbol}
\usepackage{cite}
\usepackage{wrapfig}
\usepackage{units}
\usepackage{xfrac}
\usepackage{algpseudocode}
\usepackage{algorithm}
\usepackage{bm}  
\usepackage{mathtools,amsmath}

\usepackage{hyperref}

\usepackage{tikz}
\newcommand\copyrighttext{%
  \footnotesize Accepted for presentation in IROS 2017, Vancouver, Canada. \copyright 2017 IEEE. Personal use of this material is permitted. Permission from IEEE must be obtained for all other uses, in any current or future media, including reprinting/republishing this material for advertising or promotional purposes, creating new collective works, for resale or redistribution to servers or lists, or reuse of any copyrighted component of this work in other works.}
\newcommand\copyrightnotice{%
\begin{tikzpicture}[remember picture,overlay]
\node[anchor=south,yshift=10pt] at (current page.south) {\fbox{\parbox{\dimexpr\textwidth-\fboxsep-\fboxrule\relax}{\copyrighttext}}};
\end{tikzpicture}%
}

\begin{document}

\title{\LARGE \bf
RCAMP: A Resilient Communication-Aware Motion Planner \\ for Mobile Robots with Autonomous Repair of Wireless Connectivity
}
\author{ Sergio Caccamo, Ramviyas Parasuraman, Luigi Freda, Mario Gianni, Petter \"Ogren
\thanks{The authors S.Caccamo, P.\"Ogren are with the Computer Vision and Active Perception Lab., Centre for Autonomous Systems, School of Computer Science and Communication, KTH Royal Institute of Technology, Sweden. R.Parasuraman is with Purdue University, West Lafayette, USA. L.Freda and M.Gianni are with ALCOR Laboratory, DIAG, Sapienza University of Rome, Italy.
e-mail: \tt{ $\{$caccamo$|$petter$\}$@kth.se, ramviyas@purdue.edu, $\{$freda$|$gianni$\}$@dis.uniroma1.it} }}

\maketitle
\thispagestyle{empty}
\pagestyle{empty}
\begin{abstract}

Mobile robots, be it autonomous or teleoperated, require stable communication with the base station to exchange valuable information. 
Given the stochastic elements in radio signal propagation, such as shadowing and fading, and the possibilities of unpredictable events or hardware failures, communication loss often presents a significant mission risk, both in terms of probability and impact, especially in Urban Search and Rescue (USAR) operations. 
Depending on the circumstances, disconnected robots are either abandoned, or attempt to autonomously back-trace their way to the base station. 
Although recent results in Communication-Aware Motion Planning can be used to effectively manage connectivity with robots, there are no results focusing on  autonomously re-establishing the wireless connectivity of a mobile robot 
without back-tracing or using detailed a priori information of the network.
 
In this paper, we present a robust and online radio signal mapping method using Gaussian Random Fields, and propose a Resilient Communication-Aware Motion Planner (RCAMP) that integrates the above signal mapping framework with a motion planner.
RCAMP considers both the environment and the physical constraints of the robot, based on the available sensory information. We also propose a self-repair strategy using RCMAP, that takes both connectivity and the goal position into account when driving to a connection-safe position in the event of a communication loss.
We demonstrate the proposed planner in a set of realistic simulations 
of an exploration task in single or multi-channel communication scenarios.

\end{abstract}
\begin{keywords}
\textit{Mobile Robots, Self-Repair, Wireless Communication, Communication-Aware Motion Planning.}
\end{keywords}

 \copyrightnotice

\section{Introduction}
\label{sec:intro}

\begin{figure}[t]
  \center
     \includegraphics[width=0.9\columnwidth]{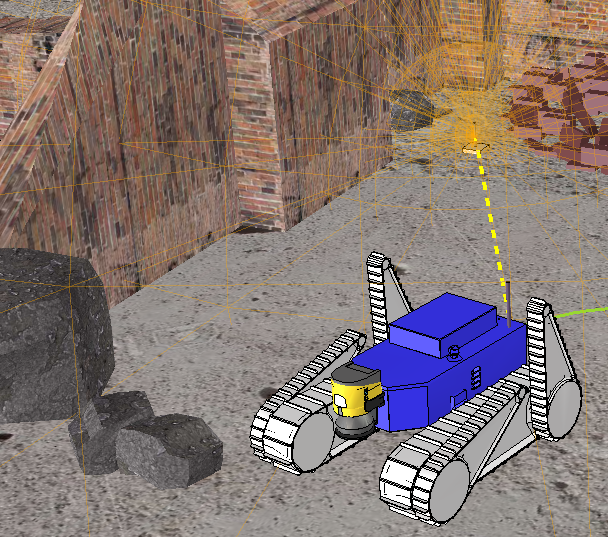}
  \caption{\small{The simulated mobile robot (UGV) with its receiver and an omnidirectional transmitter on a urban search and rescue scenario.}}
  \label{fig:setup_ugv}
\end{figure}

Recent years have witnessed an increased development of wireless technologies and significant improvements in communication performance and quality. As wireless networks possess many advantages over a tethered connection, such as the ease of deployment and fewer physical constraints, it has become the 'de facto' means of communication in mobile robots. However, this development has  not come without problems. A 2004 study \cite{Carlson2004} found a drastic increase in communication-related failures in robots compared to its prior in 2002. 

These problems are important under normal circumstances, but become even more significant in USAR scenarios, where electromagnetic infrastructure is often damaged. Furthermore, USAR missions often rely more on  bi-directional communication channels than other robotic applications, since the performance of a combined human-robot team is still superior compared to purely autonomous solutions in tasks such as  inspecting or assessing potentially hazardous areas \cite{Nagatani2013,Parasuraman2014thesis}.

To address this problem, several researchers have focused on Communication-Aware Motion (or path) Planning (CAMP) to simultaneously optimize motion and communication constraints and finding and executing an optimal path towards a destination \cite{Zhang2015}. In particular, Mostofi et al. laid solid foundations in this research area \cite{Ghaffarkhah2010,Mostofi2012,Yan2013}. 
It can be noted that most previous works consider either a binary or a disk based connectivity model, or an accurate communication model to optimize the robots motion and communication energy without focusing on resilience. Additionally, none of the previous works explicitly addresses the problem of efficiently re-establishing the communication in case of a connection loss. 

In this paper, we propose a Resilient Communication-Aware Motion Planner (RCAMP) that combines two key elements:
 1) a Gaussian Random Field (GRF) based probabilistic model to map the Radio Signal Strength (RSS) of an unknown environment and use it to predict the communication quality of the planned path; 2) a motion planning strategy that starting from sensory information (such as LIDAR), computes a traversability map for a given robot  taking into account environmental constraints. 
Additionally we propose a strategy to autonomously repair a communication loss by steering the robot towards a communication-safe position using the proposed RCAMP. 

Specifically, inspired by \cite{Fink2010}, we use GRFs for dynamically mapping the heterogeneous distribution of the RSS. We then merge this online framework with a motion planner 
\begin{itemize}
\item to obtain a semi-optimal path considering both communication and motion constraints, and 
\item to quickly re-establish connection in case of signal loss.
\end{itemize}

We demonstrate the feasibility of our approach through extensive simulations on a variety of use cases that reproduce realistic wireless network changes (e.g. a sudden connection loss) in single and multi-channel set-ups. 
The main advantages of our planner compared to others are the response to dynamic changes in the network configuration (e.g. disruptions or movement in Access Points) or in the environment (e.g. path planning in presence of dynamic obstacles) and 
the fact that we do not require prior knowledge of the network, such as the location of the Access Points. We propose a fully online, dynamic and reactive CAMP that adapts to the recent sensory information.
\section{Related Work}
\label{sec:relatedWork}

Considerable efforts have been made to address the problem of maintaining robust  wireless communication between  mobile robot(s) and a base station \cite{Rizzo2016,hsieh2008,Parasuraman2014thesis}. Many solutions focus on using mobile repeater (relay) robots to establish and/or repair an end-to-end communication link \cite{Min2016,Parasuraman2014b,kim2013mobile}. Other solutions focus on means to provide situation awareness of wireless connectivity to the robot or the teleoperator \cite{Caccamo2015}.
 
An overview of the CAMP problem is presented in \cite{Zhang2015}. Several works rely upon an oversimplified model in which the connectivity is modelled as a binary function. In this case, the predicted Signal to Noise Ratio (SNR) and the estimated distance from the robot (aerial or ground) to the radio source are empirically thresholded in order to identify regions with high probability of communication coverage  \cite{johansen2012task}. 
 
In \cite{DiCaro2014}, the authors propose an optimization strategy to compute a path along which the predicted communication quality is maximized. They make use of supervised learning techniques (Support Vector Regression) to predict the link quality such as the Packet Reception Ratio. It is worth noting that in this case the learning mechanism is offline and hence  can only be applied to a static environment. 
 
A communication aware path planner is proposed in \cite{stachura2011} for an aerial robot. Here, the authors present a probability function which is based on the SNR between two nodes. The SNR model is learned from the measurements online using an Unscented Kalman Filter (UKF) model. 
 
Works that combine communication and motion planning  are strongly influenced by Mostofi et al. In \cite{Mostofi2012}, the authors developed a mathematical framework to predict the communication quality (mainly the SNR) in unvisited locations by learning the wireless channels online. This prediction model is then used to define a motion planner either to improve the channel assessment \cite{Ghaffarkhah2010} or to optimize for communication and motion energy to reach a given target \cite{Yan2013}. This framework is further extended in \cite{ali2016motion} to include online channel learning for co-optimization of communication transmission energy and motion energy costs. Here, the transmission power is modelled as a function of SNR, whereas the motion power is a function of the robot's velocity and acceleration.
 
Recovering  from a communication failure is a topic that has not been given much attention in  the community. A simplistic solution is to back-track the robot along the path it has already travelled, until it regains communication. Alternatively, the robot can predict positions where the connection has high quality and move towards those locations in case of connection loss. In \cite{Derbakova2011}, a decentralized algorithm is proposed to move the disconnected robot towards the known position of the gateway (radio signal source or relay) by taking into account obstacles along the way. In \cite{hsieh2008}, the authors demonstrated a behaviour to drive the disconnected robot towards the closest robot node (assuming a multi-robot network) and repeat this until connection is restored. Note that in the above mentioned works, the wireless channel parameters are not estimated, but instead  perfect knowledge on the network topology is assumed (e.g, the positions of the gateway nodes, base station, etc.).
 
In the Wireless Sensor Networks (WSN) community, where it is commonly assumed that ample amounts of hopping nodes are available, the problem of repairing a connectivity failure is viewed differently. In this case, mobile robots can be used as sensor nodes which can be repositioned or added to replace failed nodes \cite{wu2011wireless,Truong2016}. 
 
It can be seen that predicting the communication quality in regions not explored by a mobile robot is a challenging problem. As pointed out above, probabilistic approaches such as maximum likelihood and UKF have been used to model the path loss and shadowing components of the RSS. Yet these models perform efficiently only when there is at least some prior information available regarding the network, such as source or relay node positions, which is difficult to know in field robotics applications such as the emergency deployment of robots to help in disaster response operations. In \cite{Fink2010}, a Gaussian Process based method is proposed to estimate the channel parameters and map the RSS in real-time using a few sample measurements. Taking inspirations from this work, in this paper, we propose a truly online Gaussian Random Field model to assess the RSS by continuously learning from the field measurements. 
 
We make use of this probabilistic model to obtain the communication cost of a given path. We then co-optimize this cost along with the motion costs (ensuring feasibility of traversal by taking into account environment obstacles and constraints) to compute a path to a given destination. The motion planner then executes this path by actively re-planning. In case of a connection loss and if no destination is defined, the motion planner makes use of the online GRF model to quickly drive to a position that has the highest probability to restore connectivity, by setting the robot's starting position as the goal.

\section{Methodology}
\label{sec:Method}

In this section, we first define the RSS model, and then discuss how to apply Gaussian Random Fields (GRF\footnote{GRF is a term for the Gaussian Process Regression with 2.5 dimensional datasets where each $x-y$ coordinate has a single value $v$.}) to generate an online prediction map of the RSS distribution which will be used in both motion planning and reconnection planning. We conclude this section with a description of the Communication-Aware Motion Planner and its utility function. Note that the method can be extended to 3D and hence be applied to aerial robots as well. 

\subsection{Radio Signal Strength Model}
\label{sec:rssmodel}
When a radio signal propagates from a source to a destination, its strength attenuation depends on environmental factors such as distance (path loss), objects in the environment (shadowing) and spatio-temporal dynamics (multipath fading) \cite{Lindhe2007}. A frequently used model to represent the RSS is given by \cite{Rappaport2001}:
\begin{align}
RSS_{(d,t)} = \underbrace{RSS_{d_0} - 10\eta\log_{10}(\frac{d}{d_0})}_{path\;loss} - \underbrace{\Psi_{(d)}}_{shadowing} - \underbrace{\Omega_{(d,t)}}_{multipath} .
\label{eqn:elnsm}
\end{align}
Here, $RSS_{d_0}$ is the RSS at a reference distance $d_0$ (usually 1m), which depends on the transmit power, antenna gain, and the radio frequency used. $\eta$ is the path loss exponent which is a propagation constant of a given environment. $d = ||x - x_0||$ is the distance of the receiver (at position $x$) from the radio source (at position $x_0$). $\Psi \sim \mathcal{N}(0,\sigma)$ is a Gaussian random variable typically used to represent shadowing while $\Omega$ is a Nakagami-distributed variable representing multipath fading. 

Usually, the RSS measurements (in dBm) coming from wireless adapters are prone to noise and temporal fluctuations in addition to multipath fading. This noise can be mitigated by applying an exponentially weighted moving average (EWMA) filter~\cite{Parasuraman2014b}: 
\begin{equation}
{RSS^f(i)} = RSS^f(i-1) + \alpha (RSS(i)- RSS^f(i-1)),
\end{equation}
where $RSS(i)$ is the RSS value measured at the $i^{th}$ instant, $RSS^f$ is the filtered RSS value and $\alpha$ is an empirical smoothing parameter. 

We use  Gaussian Processes for regression (GPR) \cite{rasmussengp} for modeling the radio signal distribution as  demonstrated in \cite{Fink2010,muppirisetty2016spatial,Ferris2007}. A key difference compared to the previous approaches is that we employ online learning with dynamic training size that adapts to the changes in the environment (e.g. change from line of sight to non-line of sight of the source, switching between access points, losing/regaining a connection, etc.). Below we briefly describe how the GPR is performed.

\subsection{Gaussian Random Fields}
\label{sec:gaussianprocessesrf}

The RSS distribution can be described with a function $f:\mathbb{R}^2\to \mathbb{R}$ where each vector of \textit{xy}-coordinates generates a single RSS.
Such a function can be efficiently modeled by a GRF which places a multivariate Gaussian distribution over the space of $f(\mathbf{x})$.
The GRF allows us to probabilistically handle noisy meausurements of a dynamic and unknow process and predict the behaviour of such a process at unknown and unexplored states. 
GRF have been widely used on a broad range of robotics problems such as haptic and visual perception \cite{caccamogpr}, geometric shape description and planning \cite{dragievgp}.
As shown in \cite{Ferris2007}, environmental observation of RSS can condition a GRF so that its posterior mean defines the signal distribution of interest.
The GRF is in fact shaped by a mean function $m\left(\mathbf{x}\right)$ and a covariance function $k\left(\mathbf{x_i},\mathbf{x_j}\right)$.

To properly describe the probabilistic model we define the set $\textit{R}_{V} = \{\mathbf{r_1},\mathbf{r_2}\dotsc\mathbf{r_N}\}$, with $\mathbf{r_i} \in \mathbb{R}^3$, of measurements of robot xy-positions and corresponding RSS. $D_{RF} = \{\mathbf{x_i},y_i \}_{i=1}^N $ is a training set where $\mathbf{x_i}\in \mathbf{X}\subset\mathbb{R}^2$ are the \textit{xy}-coordinates of the points in $\textit{R}_{V}$ and $y_i$ the RSS readings from the mobile robots wireless adapters.  $\mathbf{X_*} \equiv \mathbf{X_{{rf}_*}} \subset\mathbb{R}^2$ represents a set of $M$ test points where $\mathbf{x_{{rf}_i}} \in \mathbb{R}^2$ is a xy-coordinate of the environment.

 The joint Gaussian distribution on the test set $\mathbf{X_*}$, assuming noisy observation $ \mathbf{y} = f\left(\mathbf{x}\right) + \epsilon  \text{ with } \epsilon \sim \mathcal{N}\left( 0, \sigma_n^2\right)$, assumes the following form

\begin{equation}
\begin{bmatrix}
  \mathbf{y} \\
  \mathbf{f_*}
\end{bmatrix}
\sim \mathcal{N}\left( m\left(\mathbf{x}\right),
\begin{bmatrix}
  \mathbf{K} + \sigma_n^2I & \mathbf{k_*}\\
  \mathbf{k_*^T} & \mathbf{k_{**}}
\end{bmatrix}
\right)
\label{eqn:predictivefunct}
\end{equation}
where $\mathbf{K}$ is the covariance matrix between the training points $\left[\mathbf{K}\right]_{i,j = 1 \dotsc N} = k\left(\mathbf{x_i},\mathbf{x_j}\right)$, $\mathbf{k_*}$ the covariance matrix between training and test points $\left[\mathbf{k_*}\right]_{i=1 \dotsc N,j=1 \dotsc M} = k\left(\mathbf{x_i},\mathbf{{x_*}_j}\right)$ and $\mathbf{k_{**}}$ the covariance matrix between the only test points $\left[\mathbf{k_{**}}\right]_{i,j=1 \dotsc M} = k\left(\mathbf{{x_*}_i},\mathbf{{x_*}_j}\right)$.

We use the popular squared-exponential kernel
\begin{equation}
k\left(\mathbf{x_i},\mathbf{x_j}\right) = \sigma_e^2 \text{exp}\left( - \frac{\left(\mathbf{x_i}-\mathbf{x_j}\right)^T\left(\mathbf{x_i}-\mathbf{x_j}\right)}{\sigma_w^2}\right).
\label{eqn:kernelexp}
\end{equation}
as it better represent the variance in RSS \cite{Fink2010,Ferris2007}. 

Following the example of \cite{Fink2010}, we could define a model-based potential prior based on the path loss eq.~(\ref{eqn:elnsm}) to improve the accuracy of prediction
\begin{equation}
m\left(\mathbf{x}\right) =  RSS_0 -10 \eta \log_{10} \left( \Vert x -x^{s} \Vert \right) ,
\label{eqn:modelbasedprior}
\end{equation}
where $x^s$ is the source location which is an unknown parameter in the mean function. One could potentially optimize the mean hyper-parameters ($\theta_m = [RSS_0, \eta, x^s]$) by training the model with the measured data. In \cite{Fink2010,muppirisetty2016spatial,Ferris2007}, they either assumed the knowledge of the source location or estimated it  in a dedicated control/training phase with the measured data. 

However, given the unbounded nature of the source location $x^s$ and the fact that only sparse measurements in a limited explored area is available in a practical robotic application, optimizing these hyper-parameters will result in extensive computation and low accuracy. 
 
Moreover, this model can be applied only to a fixed radio source (Access point). Therefore, considering a practical USAR scenario, where the Access Points can be mobile or is frequently moved, trying to optimize the source location in eq.~(\ref{eqn:modelbasedprior}) with the measured data will not only be inaccurate, but also result in poor prediction performance of the GPR model.

Finally, more complex potential priors can be used or interchanged in order to incorporate propagation phenomenas (e.g. attenuation due to walls, floors, etc.) or environmental knowledge and improve the prediction on those regions of the map far from the measured data \cite{Strom2012}.  However, such approaches require a larger amount of information and increase the number of hyperparameters to be optimized.

Thus in our work, we consider a constant mean function,
\begin{equation}
m\left(\mathbf{x}\right) =  C,
\label{eqn:meanfunc}
\end{equation}
for practical and computational aspects. Note that this mean function has shown low prediction errors in \cite{richter2015revisiting} when compared to a linear mean function.

The predictions are obtained from the GPR conditioning the model on the training set \cite{rasmussengp} :
\begin{equation}
p\left(f_* | \mathbf{X_*},\mathbf{X},\mathbf{y}\right) =  \mathcal{N}\left(\overline{f_*},\mathbb{V}\left[f_*\right]\right)
\label{eqn:predictivefunctcond}
\end{equation}

\begin{equation}
\overline{f_*} = m\left(\mathbf{x}\right) + \mathbf{k_*^T}\left(\mathbf{K} + \sigma_n^2\mathbf{I}\right)^{-1} \left( \mathbf{y} -m\left(\mathbf{x}\right) \right)
\label{eqn:eqmeangp}
\end{equation}
\begin{equation}
V\left[f_*\right] = \mathbf{k_{**}}-\mathbf{k_*^T}\left(\mathbf{K} + \sigma_n^2\mathbf{I}\right)^{-1}\mathbf{k_*}
\label{eqn:eqsdgp}
\end{equation}

The predictive variance of the GRF highlights regions of low density or highly noisy data. 
The hyper-parameters of the mean and the kernel $\theta=[C,\sigma_c,\sigma_w]$ are periodically optimized while the mobile robot moves and collects measurements. The optimization (hyperparameter estimation) is done by maximizing the marginal logarithmic likelihood of the distribution on the measured data. 

For online optimization purposes, we efficiently train the GPR after each measurement by dynamically adjusting the training set size based on the magnitude of the changes in the measurements. We optimize the GPR and start with the RSS prediction after the robot has moved enough to acquire the minimum amount of training samples (around 5 meters of displacement). The GPR model is continuously re-trained with every new collected sample. When the connection status is active, we keep increasing the training set size up to a certain maximum limit. If the connection is lost, we keep decreasing the training size until the minimum limit. The hyper-parameters are re-optimized with current measurements whenever the training  size reaches a certain minimum.

Next we describe how to define a utility function that takes into account the prediction of the RSS and its uncertainty to generate trajectories that guarantee communication or to re-establish a connection in case of signal loss.
We use the term "Wireless Map Generator (WMG)" to refer the above described Gaussian random field that generalizes over robot positions and RSS measurements to generate wireless distribution maps.

\subsection{Communication-Aware Motion Planner}
\label{sec:communicationawaremotionplanner}

We use the RSS predictions from the GPR along with the traversibility cost in the RCAMP to plan and execute a path to a given destination. As the planner is dynamic, it keeps track of both RSS predictions and the traversibility based on the incoming sensory information. We detail the basic steps below. 

\subsubsection{Mapping and Point Cloud Segmentation}

As a necessary prerequisite for path planning, a \emph{map} representation $\cal M$ of the environment is incrementally built in the form of a point cloud. An ICP-based SLAM algorithm is used in order to register the different 3D laser scans collected by the robot. At each new scan, both the map and a structure interpretation of it are updated. In particular, the point cloud map $\cal M$ is segmented in order to estimate the traversability of the terrain. 

In a first step, $\cal M$ is filtered using an efficient occupancy voxel-map representation~\cite{hornung2013octomap}: recursive binary Bayes filtering and suitable clamping policies ensure adaptability to possible dynamic changes in the environment. 

Next, geometric features such as surface normals and principal curvatures are computed and organized in histogram distributions. Clustering is applied on 3D-coordinates of points, mean surface curvatures and normal directions~\cite{menna2014real}. As a result, a classification of the map $\cal M$ in regions such as \textit{walls}, \textit{terrain}, \textit{surmountable obstacles} and \textit{stairs/ramps} is obtained. 

\subsubsection{Traversability Cost}
Traversability is then computed on the map $\cal M$ as a cost function taking into account point cloud classification and local geometric features~\cite{ferri2014point}. In particular, the traversability cost function $trav:\mathbb{R}^3\longmapsto\mathbb{R}$
is defined as
\begin{equation}
trav(\mathbf{p}) = w_L(\mathbf{p})(w_{Cl}(\mathbf{p}) + w_{Dn}(\mathbf{p}) + w_{Rg}(\mathbf{p}))
\end{equation}
where $\mathbf{p} \in \mathbb{R}^3$ is a map point, the weight $w_L(\mathbf{p})$ depends on the point classification, $w_{Cl}(\mathbf{p})$ is a function of the robot obstacle clearance, $w_{Dn}(\mathbf{p})$ depends on the local point cloud density and $w_{Rg}(\mathbf{p})$ measures the terrain roughness (average distance of outlier points from a local fitting plane). 
A \textit{traversable map} ${\cal M}_t$ is obtained from ${\cal M}$ by suitably thresholding the obstacle clearance $w_{Cl}(\cdot)$ and collecting the resulting points along with their traversability cost.  

\subsubsection{Global and Local Path Planners}
Path planning is performed both on global and local scales. 
Given a set of waypoints as input, the \textit{global} path planner is in charge of (1) checking the existence of a traversable path joining them and (2) minimizing a combined RSS-traversability cost along the computed path. Once a solution is found, the \textit{local} path planner safely drives the robot towards the closest waypoint by continuously replanning a feasible path in a local neighbourhood of the current robot position. This allows us to take into account possible dynamic changes in the environment and local RSS reconfigurations. 

Both the global and the local path planners capture the connectivity of the traversable terrain by using a sampling-based approach. A tree is directly expanded on the traversability map ${\cal M}_t$ by using a randomized A* approach along the lines of~\cite{ferri2014point}. The tree is rooted at the starting robot position. Visited nodes are efficiently stored in a kd-tree. The current node $\bm{n}$ is expanded as follows: first, the robot clearance $\delta(\bm{n})$ is computed at $n$; second, a neighbourhood ${\cal N}(\bm{n})$ of points is built by collecting all the points of ${\cal M}_t$ which falls in a ball of radius $\delta(\bm{n})$ centred at $\bm{n}$. Then, new children nodes are extracted with a probability inversely proportional to the traversability cost. This biases the tree expansion towards more traversable and safe regions. The total traversal cost of each generated child is evaluated by using eqn.~(\ref{eq:costfunction}) and pushed in a priority queue $\cal Q$. The child in $\cal Q$ with the least cost is  selected as next node to expand.

\subsubsection{Cost Function}
The randomized A* algorithm computes a sub-optimal path $\{\bm{n}_i\}_{i=0}^N$ in the configuration space ${\cal C}$ by minimizing the total cost 
\begin{equation}
J = \sum_{i=0}^N c(\bm{n}_{i-1},\bm{n}_i) \, ,
\end{equation} 
where $\bm{n}_0$ and $\bm{n}_N$ are respectively the start and the goal configurations, and  ${ \bm{n}_i \in \cal C}$. 
In this paper we define the cost function $c:{\cal C}\times{\cal C}\longmapsto \mathbb{R}$ so as to combine  traversability and RSS predictions. In particular
\begin{equation}\label{eq:costfunction}
\begin{multlined}
c(\bm{n}_i,\bm{n}_{i+1}) = \big(d(\bm{n}_i,\bm{n}_{i+1}) + \\  
h(\bm{n}_{i+1}, \bm{n}_N)\big)\pi_1(\bm{n}_{i+1}) \pi_2(\bm{n}_{i+1}) 
\\
\pi_1(\bm{n}) = \lambda_t \frac{trav(\bm{n})-trav_{min}}{trav_{max}-trav_{min}+\varepsilon} +1 \\
\pi_2(\bm{n}) = \lambda_r \alpha_r e^{-t/\tau} \frac{ rss_{max} - rss(\bm{n})}{rss_{max}-rss_{min}+\varepsilon} +1 
\end{multlined} 
\end{equation}
where $d:{\cal C}\times{\cal C}\longmapsto \mathbb{R}^+$  is a distance metric, $h:{\cal C}\times{\cal C}\longmapsto \mathbb{R}^+$ is a goal heuristic, $\lambda_t, \lambda_r \in  \mathbb{R}^+$ are scalar positive weights, $rss:{\cal C}\times{\cal C}\longmapsto \mathbb{R}$ is the estimated RSS, $\alpha_r \in \left[ 0,1 \right] $ is a confidence which can be obtained by normalizing the variance of the RSS prediction (as returned by the GPR), $\varepsilon$ is a small quantity which prevents division by zero and $\tau$ is an exponential decay constant (determines the amount of time after which $\pi_2$ goes to its minimum value 1). In particular, with abuse of notation we use $trav(\bm{n})$ to denote the traversability of the the point corresponding to $\bm{n}$. The first factor in eq.~(\ref{eq:costfunction}) sums together the distance metric and the heuristic function (which depends on the distance to the goal). The other two factors $\pi_1$ and $\pi_2$ respectively represent a normalized traversability cost and a normalized RSS cost, whose strengths can be increased by using the weights $\lambda_t$ and $\lambda_r$ respectively ($\pi_i \geq 1$). The exponential decay is used to decrease the effect of the RSS cost after a certain time (e.g. before the path planner is stopped by a timeout in case a path solution is difficult to  find). 

Note, instead of jointly optimizing the motion and communication energy for a given path as in \cite{Yan2013}, we plan an optimized trajectory to a given goal position using a cost function  that represents a balanced optimization between communication and traversibility costs, includes normalization of the used metrics, and allows setting different priorities using the parameters $\lambda_t$ and $\lambda_r$. 

\paragraph*{Self-recovery} 
The cost function in eq.~(\ref{eq:costfunction}) gives us the leverage in generating a trajectory that recovers from communication loss. In the case of a connection loss, we define the goal position as the robot's initial position or the AP position (if known), so as to bound the  search and to guarantee the re-establishment of connectivity.

\section{Experimental evaluation}
\label{sec:Experiments}

\begin{figure} 
  \center
     \includegraphics[width=\columnwidth]{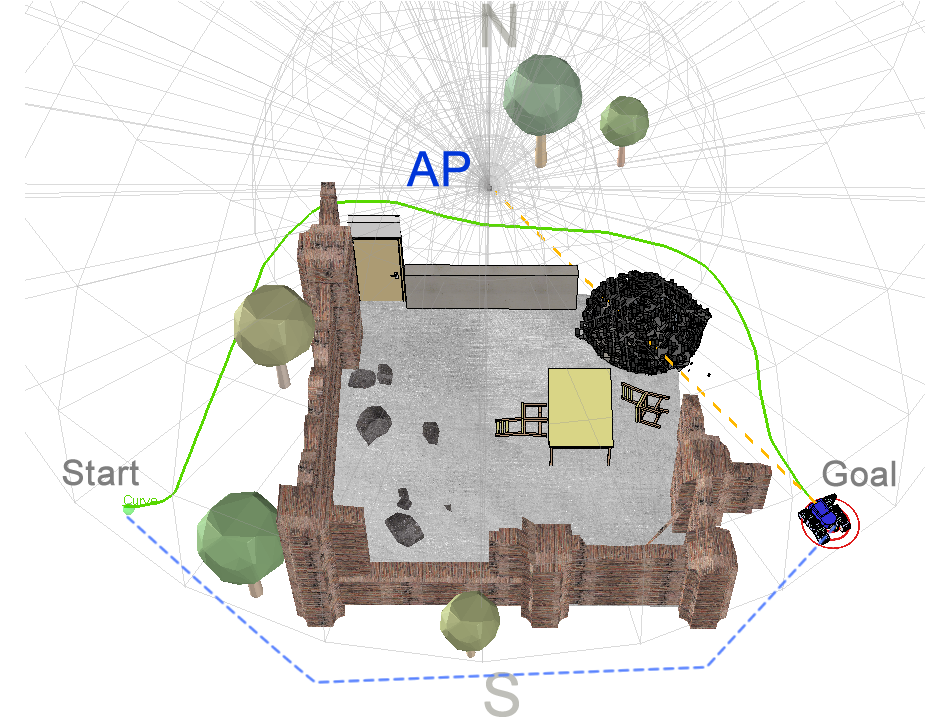}
  \caption{\small{Experimental scenario 1. The UGV tries to reach the goal position avoiding connection drops. The blue dotted line represents the shortest path, that will cause   a connection loss (going outside the AP range). The green line represents a path that reaches the goal position while keeping the robot connected to the AP. }}
  \label{fig:pic_experiment1}
\end{figure}
 
We evaluated the performance of the proposed method through a series of experiments made on simulations using V-REP. Using the 3D model of the real UGV used in \cite{tradr} we created 3 different simulation environments, reproducing typical USAR use cases, containing several obstacles and sources of signal (APs). The AP is simulated following eq.~(\ref{eqn:elnsm}) with typical parameters such as $\eta =3$, $\sigma=2$ \cite{Rappaport2001} considering a 2.4 GHz Wi-Fi communication. For each environment, we changed the positions of the robot and APs and repeated the experiments in several trials. All the software components including the RCAMP ran under the Robot Operating System (ROS). 

Note we do not evaluate the GRF model separately. Nevertheless, the GRF with mean functions in eq.~(\ref{eqn:modelbasedprior}) and (\ref{eqn:meanfunc}) have shown to perform well in signal source prediction and location estimations \cite{muppirisetty2016spatial,Ferris2007,richter2015revisiting}.
 
\subsection{Experimental scenarios}
\label{sec:expscenario}
 
\textit{Scenario 1:} In the first scenario, see Fig. \ref{fig:pic_experiment1}, the UGV is placed on the start position and must traverse an area containing a damaged building, to reach the goal position. An AP is placed on the northern part of the map (zone N in Fig. \ref{fig:pic_experiment1}). The AP uses an omni-directional antenna covering a circular area that extends to half of the map, leaving  the southern part (zone S in Fig. \ref{fig:pic_experiment1}) uncovered.  Start and goal positions are placed such that the shortest connecting path between the two points would traverse the poorly connected part of the map (S). Thus, RCAMP must generate a trajectory that connects the start and goal positions while keeping the robot in the signal covered area avoiding communication drops. With this scenario we want to demonstrate the capability of our utility function in keeping the robot connected to the AP.

\begin{figure}[t]
  \center
     \includegraphics[width=\columnwidth]{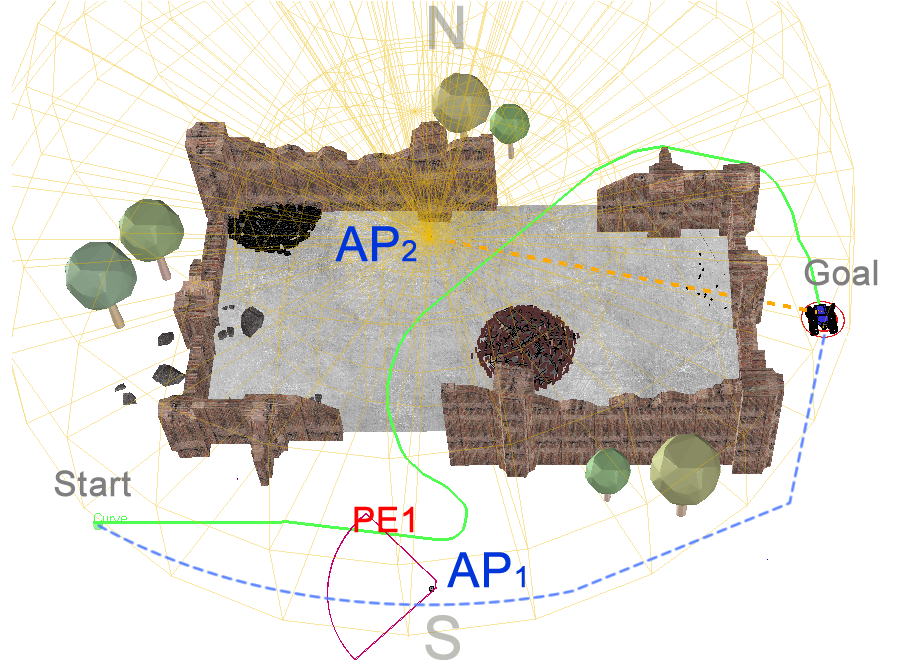}
  \caption{\small{Experimental scenario 2. The UGV tries to reach the goal position avoiding connection drops. The blue dotted line represents the shortest path to the goal position. The UGV is connected to  AP$_1$ in the first part of the path. PE1 indicates the location of the UGV when AP$_1$ shuts down after a simulated hardware failure. The green line represents a new path that reaches the goal position while keeping the UGV connected, after switching from AP$_1$ to AP$_2$.}}
  \label{fig:pic_experiment2}
\end{figure}

\textit{Scenario 2:} In the second scenario, see Fig. \ref{fig:pic_experiment2}, two different APs cover the whole map.
In this use case we want to test the promptness of the RCAMP to adapt to drastic changes in the wireless signal distribution.
The robot starts the mission connected to AP$_1$. The RCAMP must generate a path from the start position to the goal position that ensures WiFi coverage. During the mission, AP$_1$ is switched off when the robot enters the region PE1, so to simulate a communication loss event. When the connection is lost, the robot  connects to other APs (if available) in the same network, in a typical roaming behaviour. Once the robot connects to AP$_2$, the WMG must adapt its predictive model to the new signal distribution accordingly and reshape the RSS map. The RCAMP must then promptly update the path to the goal to ensure WiFi coverage.
 
\begin{figure}
  \center
     \includegraphics[width=\columnwidth]{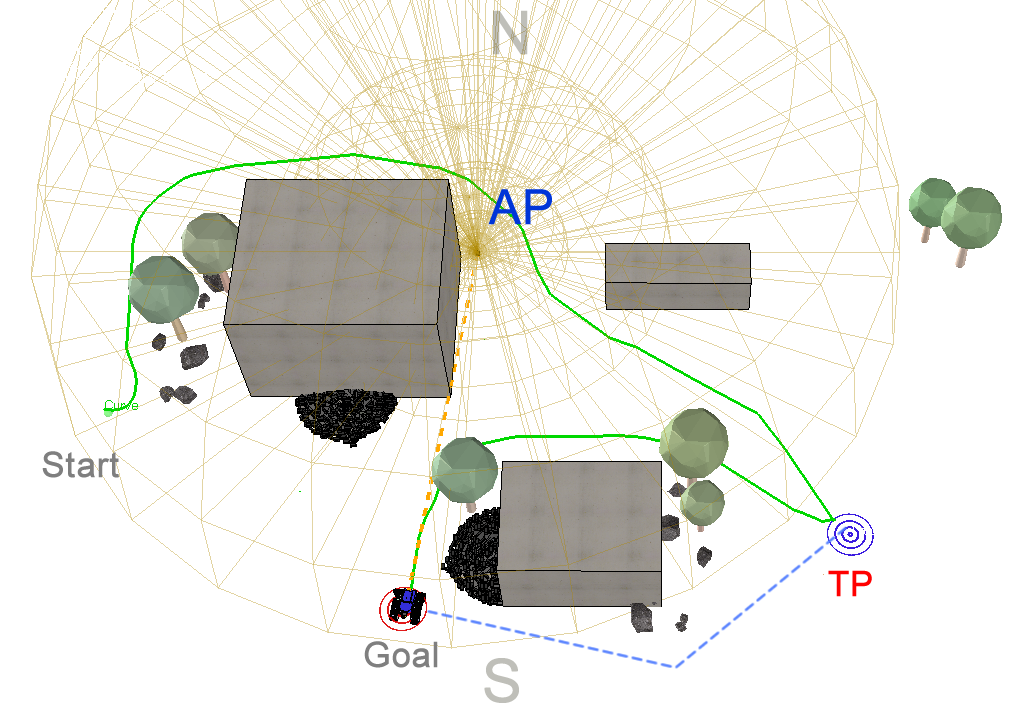}
  \caption{\small{Experimental scenario 3. The UGV is teleoperated in a USAR mission. The operator drives the robot outside the WiFi coverage area (at point TP) and the connection is lost. The system  autonomously re-establishes the connection driving the UGV to a location with high RSS and then continues to reach the goal.}}
  \label{fig:pic_experiment3}
\end{figure}

\textit{Scenario 3:} Finally, in the last scenario, see Fig. \ref{fig:pic_experiment3}, we test our self-repair strategy in case of a complete connection loss event. The UGV is tele-operated until the connection drops (blue circle, outside the WiFi coverage area). The goal position (red circle) cannot be reached with teleoperation because of the missing communication channel. In this scenario, the UGV must autonomously re-establish the connection while moving to the goal position. If the goal position was not specified (e.g. during an exploration task) the UGV must move to the closest location in the map where the RSS is high enough to ensure re-connection to the AP. 

\section{Results}
\label{sec:results}

\begin{figure*}[ht]
  \center
     \includegraphics[trim={10cm 0 6cm 0},clip,width=1.05\textwidth]{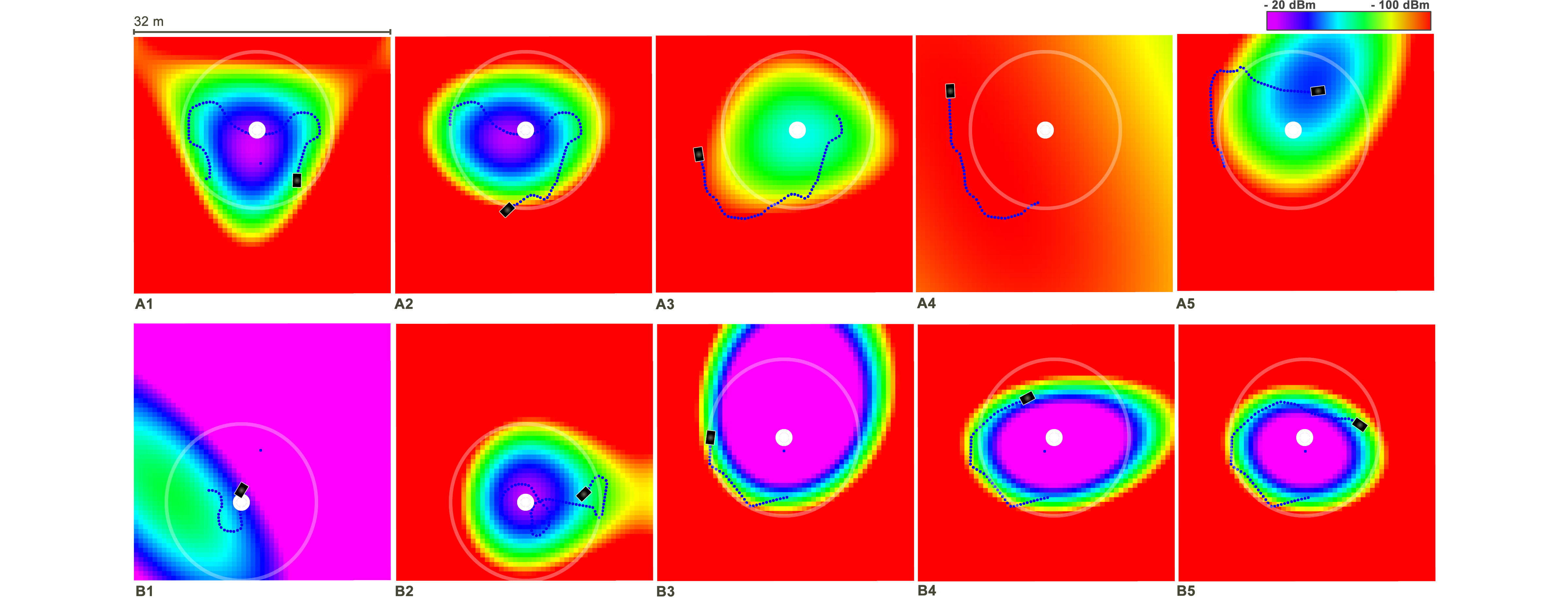}
  \caption{\small{Radio signal distributions for various trajectories in the maps of scenario 1 (A1-A5) and 2 (B1-B5). The white points represent the APs positions along with their operative ranges. The blue trajectories represent the training samples for the WMG. We can observe the changes in the RSS map generated by the WMG as the robot explores the region (without RCAMP). Note in A4 the robot is initially connected but is in a disconnected region the moat of the trajectory. }}
  \label{fig:pic_experiment2_charts}
\end{figure*}
\begin{figure}[h]
  \center
     \includegraphics[trim={2cm 2cm 0 1cm},clip,width=\columnwidth]{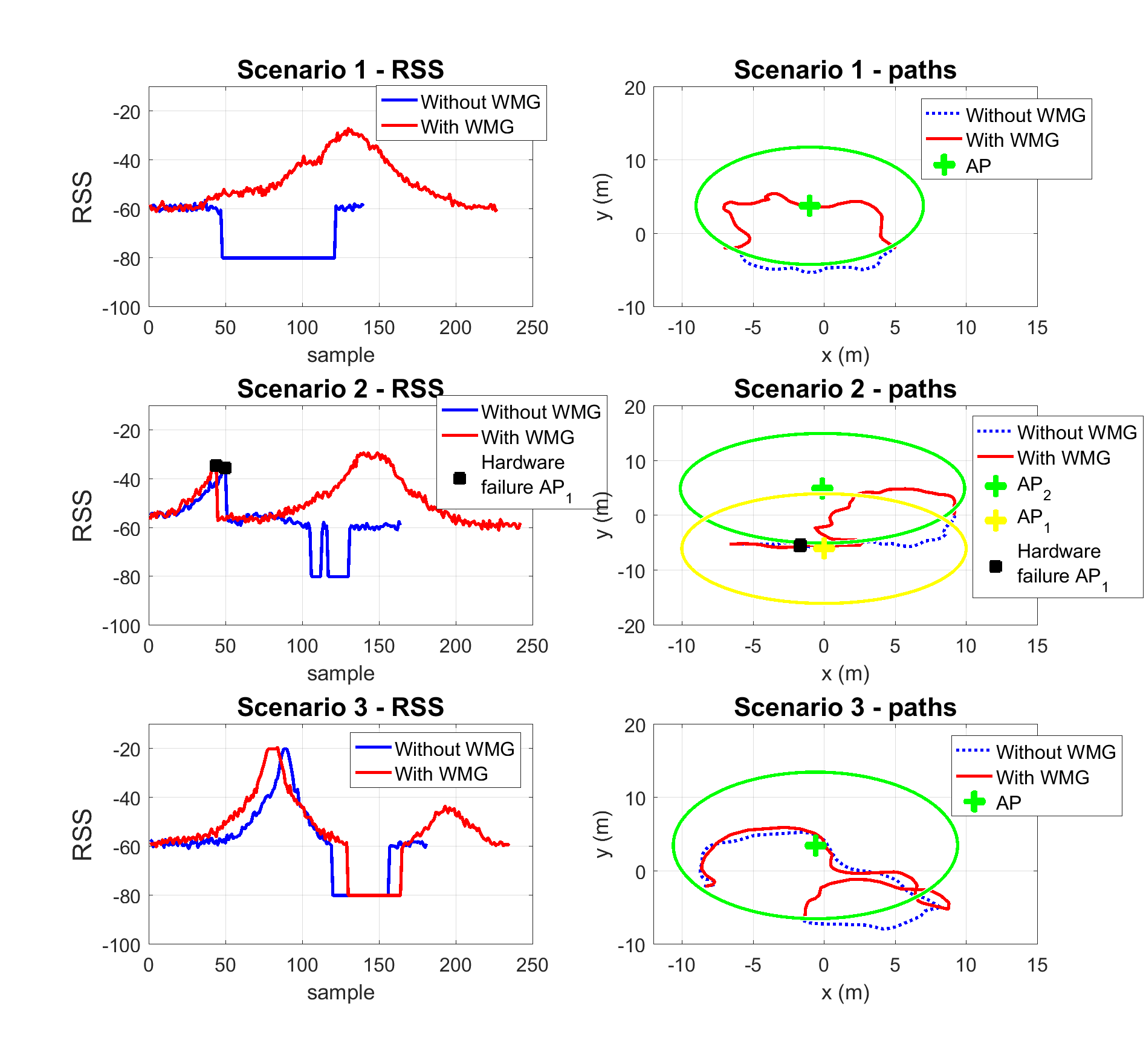}
  \caption{\small{Comparison between our RCAMP versus a normal path planner. The first column shows the RSS values measured using the on-board antenna during the three experimental scenarios. The RCAMP enables the robot to maintain an higher RSS value throughout the whole exploration. The second column shows the trajectories from start position to goal position for the three scenarios. }}
  \label{fig:pic_experiment1_charts}
\end{figure}

In the following we discuss the results of the experiments described in Sec. \ref{sec:Experiments}.
Fig.~\ref{fig:pic_experiment1_charts} shows the recorded RSS and the path taken for the three scenarios.
We present a comparison between the proposed RCAMP and a common path planner (PP). In the the first column we report the RSS values sensed by the antenna on-board the mobile robot. 

In the first row (first experimental scenario) the PP leads the robot to lose connection whereas the RCAMP defines a trajectory that maintains the robot inside the operative range of the radio transmitter as it is possible to see in the second column of the same row. The second row of Fig. \ref{fig:pic_experiment1_charts} shows that the RCAMP adapts to the drastic variation of the radio signal distribution (due to the simulated hardware failure and consecutive connection loss) and modifies the trajectory accordingly maintaining the robot inside the operative range of the new AP. The PP leads the robot to lose connection again. 
This demonstrates how the WMG promptly reacts to a connection loss in case a new source of signal is present. 

Finally, in the last row we present the results for the third scenario where the mobile robot, after a brief exploration step, is tele-operated outside the wireless range. 
The RCAMP first brings the robot back to a position where the connection can be reestablished  and then moves the robot to the goal position. The RSS value of the robot using the RCAMP, red signal in the third row, increases after the connection loss.

Fig. \ref{fig:pic_experiment2_charts} shows the predicted radio signal distribution (WMG) for experiments 1 and 2. A red color indicates low or missing signal whereas a blue-purple color indicates high signal strength.
As described in Sec. \ref{sec:gaussianprocessesrf}, the training set consists of the last visited points in the environment along with the measured RSS. The size of the training set depends on the quality of the sensed signal. The first row (A1-5) shows the predicted radio signal distribution during the first experiment. 

When the robot drives inside the operational range of the AP the training set increases and the model predicts correctly the position and the shape of the radio signal distribution (A1,2,5). Viceversa, when the mobile robot moves outside the operational range the communication with the AP drops and the training set shrinks as there is less  useful information. This strategy allows the system to promptly adapt to a new source of signal as show in the last row. Initially the system adapts to the first source of signal (AP$_1$) as is visible in B1-2. When the first AP is shut down, the systems quickly re-sizes the training set size and the WMG converges to the new signal distribution allowing to identify the position of the second AP.
\section{Conclusions}
\label{sec:conclusions}

Robots have a major potential in aiding first responders in USAR missions. In recent robot deployments, wireless networks were used in order to support mobile robot communication. This mean of communication poses several challenges, such as sudden network breakdowns and limited communication bandwidth. Based on our own experience in helping the Italian Firefighters with our UGVs and drones (under the EU-FP7 project TRADR \cite{tradr}) to assess the damages in historical buildings after the recent earthquake in Amatrice, we concluded that the inherent limitations of a wireless network can compromise the outcome of a USAR mission. Most notably, the Access Points supporting robot communication had to be regularly relocated in order to let the robot re-estabilish communication. 
 
To address some of these challenges, we proposed a Resilient Communication-Aware Motion Planner (RCAMP). Given a goal point, the RCAMP computes a trajectory by taking into account traveled distance, communication quality and environmental constraints. We used an online Gaussian Random Field to estimate the Radio Signal Strength requested by the motion planner in order to find a feasible path that takes both traversability and connectivity into account. 
We also proposed an efficient strategy to autonomously repairing a communication loss by steering the robot towards a communication-safe position computed using the RCAMP. Alternatively, if a specific destination is available, the robot plans a path that combines the objectives of reaching  the destination, and re-establishing the connection.
 
We demonstrated the proposed framework through simulations in V-REP under realistic conditions and assumptions. 
In future work, we plan to test the presented framework on real UGVs and further evaluate and analyze the performance and limits of the algorithms through more extensive field experiments.

\section*{Acknowledgments}
The authors gratefully acknowledge funding from the European Union's seventh framework program (FP7), under grant agreements FP7-ICT-609763 TRADR.

\bibliographystyle{IEEEtran}
\bibliography{InteractivePerception,FieldRobotics,WirelessRef,WifiMapping}

\end{document}